# Zebrafish Counting Using Event Stream Data

Qianghua Chen[1], Huiyu Wang[2], Li Ming[2], Ying Zhao[*]

[1,2,*]School of Electronic Information, Shanghai Dianji University, Shanghai 201306, China

[1]Email: 226003010201@st.sdju.edu.cn

[*]Corresponding author

**Abstract**

Zebrafish share a high degree of homology with human genes and are commonly used as model organism in biomedical research. For medical laboratories, counting zebrafish is a daily task. Due to the tiny size of zebrafish, manual visual counting is challenging. Existing counting methods are either not applicable to small fishes or have too many limitations. The paper proposed a zebrafish counting algorithm based on the event stream data. Firstly, an event camera is applied for data acquisition. Secondly, camera calibration and image fusion were preformed successively. Then, the trajectory information was used to improve the counting accuracy. Finally, the counting results were averaged over an empirical of period and rounded up to get the final results. To evaluate the accuracy of the algorithm, 20 zebrafish were put in a four-liter breeding tank. Among 100 counting trials, the average accuracy reached 97.95%. As compared with traditional algorithms, the proposed one offers a simpler implementation and achieves higher accuracy.

**Introduction**

Zebrafish genome is highly homologous to humans. Their tissue types are similar with those of the human body and have almost the same physiologic response to xeno-substances as mammals [1]. Because zebrafish are small, easy to breed, and have low maintenance costs, they are widely used as model organism in biomedical experiments [2]. Organism management in biomedical laboratories is highly regulated, and the number of zebrafish is a monitored metric in daily operations. Therefore, appropriate methods are required to monitor the number of zebrafish.

Existing fish counting methods can be divided into two categories: manual visual counting and computer vision-based counting. Counting fish manually is time-consuming and error-prone, and can potentially harm the fish's health[3]. Computer vision-based methods improved the automation level of aquaculture production and management[4]. At the same time, the rapid development of deep learning and neural networks has promoted the application of computer vision technology in image processing, object detection, and image segmentation[5][6]. Deep learning-based fish counting algorithms allow for rapid and precise individual fish locating, and help to quickly and effectively assessing changes in fish populations.

In computer vision counting methods, images are typically captured using optical or sonar cameras. Francisco J. Silvério et al.[7] proposed a zebrafish counting algorithm, in which an optical camera was used, and the Gaussian Mixture Model (GMM) background subtraction and blob counting techniques were combined to finish the counting. Although detection algorithms specifically designed for zebrafish are few, there are various methods for detecting fish populations. Costa et al.[8] used binocular cameras to capture underwater videos of tuna and acquired multiple biometric information to analyze the numbers of tuna. Shahrestani et al.[9] recorded 59 consecutive hours using an adaptive resolution imaging sonar, and successfully counted fish at an accuracy rate above 94%. Jing et al.[10] deployed a dual-frequency identification sonar under the water to obtain sonar images and estimate the abundance of fish. Chang et al.[11] determined the counting area by detecting the overlapping 3-dimensional point cloud regions generated from the sonar and binocular images, and then calculated the abundance of specific fish species inside the net cage. Hernan'dez-Ontiveros et al[12] employed a binarization method to detect fish body contours, extracted fish features like area, perimeter, and noises, and achieved the automatic counting of ornamental fish by setting area and perimeter thresholds.

The counting challenge for zebrafish is their small size. Smaller sizes mean that zebrafish occupy a smaller proportion of the frame, making it more challenging for machine learning models to make accurate predictions[13]. We propose a zebrafish counting algorithm that captures images using an event camera, which are extremely sensitive to moving objects. This approach leverages the moving features of zebrafish to improve counting accuracy.

**Materials and methods**

**Fish maintenance.** In this experiment, Zebrafish were maintained in self-circulating systems at Shanghai Dianji University (SDJU). This study was carried out in compliance with the ARRIVE guidelines. All methods were carried out in accordance with relevant guidelines and regulations. All experimental protocols were approved by Shanghai Dianji University.

**Event camera.** The event camera has a bio-inspired sensor. Unlike traditional cameras, the event camera doesn't provide a continuous stream of image frames at a fixed rate[14]. It only reports changes in local pixel-level brightness, thus forming an event stream. Event cameras have high temporal resolution (microseconds), high dynamic range and low power consumption. Therefore, they are ideal for challenging scenarios with traditional cameras, such as low latency, high speed, and high dynamic range. In the experiments, the event camera model is CeleX5_MP from Celepixel company with a resolution of 1280*800 pixels. The lens resolution is 10 mega pixels and the focal length is 16mm. The power consumptions in the event mode and full-picture mode are 390mW and 470mW respectively. Besides, the camera has a dynamic range of 140dB, a latency of 1 $\mu s$ /min, and an input voltage of 3.3 V.

**Preprocessing.** Different imaging methods between the RGB camera and the event camera is shown in Fig. 1A. The application scenario is an indoor environment with uniform lighting. When capturing images with an event camera, various factors may cause noises, such as the circuitry and environmental conditions. Fixed pattern noise (FPN) is the term given to a particular noise pattern on digital imaging sensors. To reduce FPN, background modeling and subtraction are applied. The steps are as follows: first, the camera lens is covered with a A4 paper; second, the camera captures the video for a few seconds and gets the background model; then, the model is subtracted from the real-time event stream to reduce the FPN(Fig. 1B). Lens distortion will make zebrafish and breeding tank out of shape. The changes in shapes may lead to misdetection and lower detection accuracy. To reduce the negative effects, Zhang's calibration[15] method is applied(Fig. 1C).

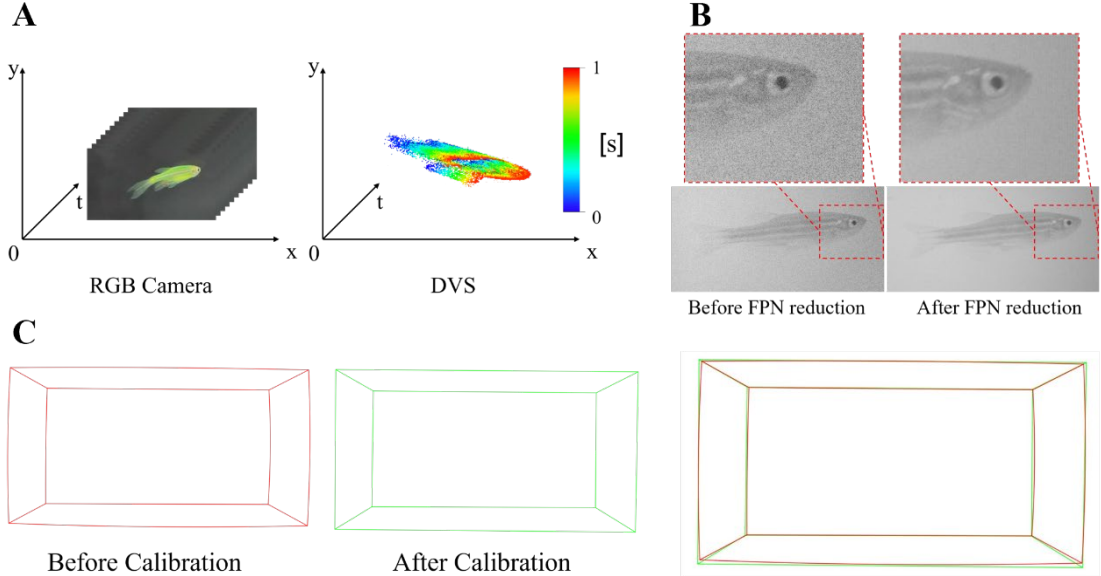

**Figure 1.** (**A**) Imaging methods between RGB and event cameras. (**B**) The comparison between an image with the FPN and the one after FPN reduction. (**C**) Red lines represent the fish tank contours before calibration, while green lines represent the contours after calibration.

**Event Images Fusion.** The event camera records data in the form of bin files, which contain event information. The event camera's SDK is used to convert the event stream data into videos in four modes, each mode can output videos or images containing different types of information. They are binary mode, count mode, gray mode, and accumulate mode respectively (Fig. 2A). Binary mode reflects the event changes, where areas with events are shown as white (RGB=255,255,255). Count mode reflects the number of times a pixel is triggered by events in a period. Gray mode reflects the grayscale value of the events. In binary, count and gray modes, areas without events are shown as black (RGB=0,0,0). In accumulate mode, the grayscale values are added over time. Changing points are replaced by new grayscale values, while unchanged points retain the grayscale value in the previous frame.

    The bin files captured by the event camera contain x, y, event on, event off, and grayscale information. When converting the bin file into images of the four modes above, some information is discarded. To keep more information for detection, the four mode images are fused together and named mixed mode. The specific fusion process is defined as Fig. 2B. In the process, a transparent channel is added to the images of binary, count and gray mode, respectively. We change the white color into green (RGB=0,255,0) to make the binary information more obvious while adding on the accumulate images. Then, the image of binary, count, gray and accumulate modes are added. In this added image, image of binary mode is on the top layer, image of count mode is on the middle layer, image of gray mode is on the low layer and image of accumulate mode is on the

lowest layer. The formula of images adding is defined as Equation (1):

$$C = 1 - (1-A) \times (1-B)$$

Where: $A$ is the color value of the up layer, $B$ is the color value of the down layer, $C$ is the resulting color value after adding.

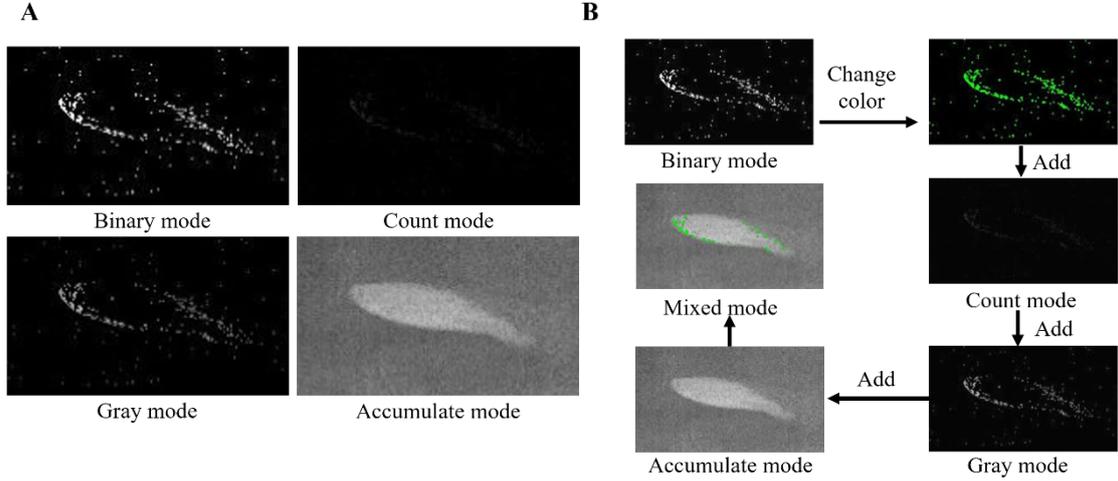

**Figure 2.** (**A**) The comparison of four modes of images provided by the event camera. (**B**) The image fusion process.

**YOLO model selection.** YOLO is a regression-based algorithm. It has been widely used due to its simple network structure and extremely high detection speed. YOLOv5 object detection model uses partial darknet and spatial pyramid pooling. The network and pooling layer compose a cross stage local network for feature extraction[16]. YOLOv5 offers various non-maximum suppression (NMS) techniques, which can help address the occlusion issue to some extent when counting them. Due to its simple architecture, fast detection, and the ability to mitigate occlusion. Functional flowchart of the network is shown in Fig. 3.

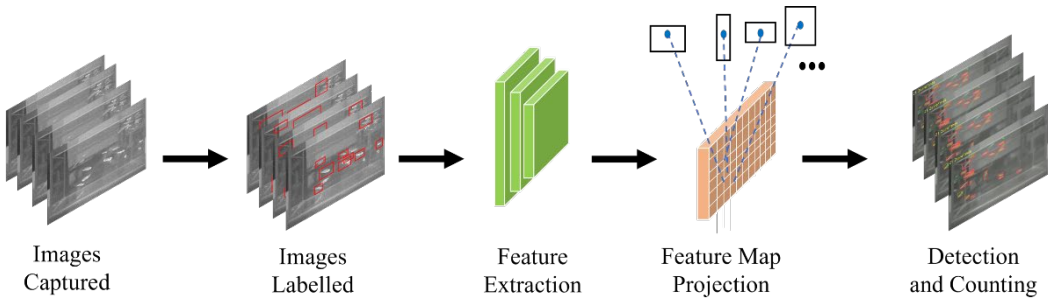

**Figure 3.** Functional network flowchart.

**Tracking.** This paper performs association and tracking of detected zebrafish. It assigns a unique ID to each detected zebrafish and helps to better count the number of zebrafish. The process of associating and tracking zebrafish is as follows:

(1) After detection, YOLO model generates a set of bounding boxes at each frame. A bounding box can be described as $[x, y, w, h]$, where $x, y$ are the coordinates of the center of the bounding box, $w, h$ are the width and height of the bounding box.

(2) Kalman Filtering is used to predict and update the state of each tracked zebrafish. The Kalman filter works by modeling the zebrafish's position and velocity to predict its future position. The position and velocity of a zebrafish are included in $x_k$, $x_k$ is defined as Equation (2):

$$x_k = [x_k, y_k, \dot{x}_k, \dot{y}_k] \tag{2}$$

where $x_k, y_k$ are the position coordinates at time step $k$ and $\dot{x}_k, \dot{y}_k$ are the velocities in the x and y directions at time step $k$. When a new detection is made, the Kalman filter predicts and updates the zebrafish's state. The Kalman filter predicts the next state $\hat{x}_{k|k-1}$ based on the previous state and the zebrafish's motion model. $\hat{x}_{k|k-1}$ is defined as Equation (3):

$$\hat{x}_{k|k-1} = F \cdot x_{k-1} + B \cdot u_k \tag{3}$$

where $F$ is the state transition matrix that models the zebrafish's motion, $B$ is the control input matrix, $u_k$ is the control vector. The predicted bounding box $\hat{b}$ can be obtained from the predicted state vector. When a new detection $z_k$ is available, the Kalman filter updates the predicted state $\hat{x}_{k|k-1}$ to a more accurate estimate $x_k$ based on the detection. $x_k$ is defined as Equation (4):

$$x_k = \hat{x}_{k|k-1} + K_k \cdot (z_k - H \cdot \hat{x}_{k|k-1}) \tag{4}$$

where $K_k$ is the Kalman gain, which determines how much weight to give to the new measurement relative to the prediction, $H$ is the measurement matrix, which relates the state vector to the observed measurement. Kalman Filtering allows the algorithm to predict the position of zebrafishes, even when they temporarily disappear due to occlusion, which helps maintain tracking continuity. In this paper, the Kalman filter is used to predict the position of the zebrafish for 15 frames.

(3) The Hungarian algorithm is used to match the detected bounding boxes in the current frame to the predicted locations. It minimizes a cost matrix based on the Euclidean distance between the predicted bounding box centers and the detected bounding boxes centers. The cost matrix is defined as Equation (5):

$$c_{ij} = \|z_i - \hat{x}_j\|_2 \tag{5}$$

where $z_i$ is the detection for zebrafish $i$, $\hat{x}_j$ is the predicted location of zebrafish $j$. The Hungarian algorithm then solves the assignment problem based on the cost matrix, finding the optimal match between detections and existing tracks.

Once the detection starts, each detected zebrafish will have a unique ID. If a zebrafish appears in the predicted trajectory, then it's ID remains the same. If a zebrafish is lost temporarily (due to occlusion), the Kalman filter's prediction helps to recover the trajectory in the next frame.

**Statistical mechanism.** To reduce the influence of the complete occlusion, a statistical mechanism is performed in the research. The process of statistical mechanism is as follows: first, three seconds of continuous mixed mode video is generated; second, and for each frame, the average counting result within this 3-second period is calculated; then, the average counting result is rounded up to the nearest integer as the final counting result.

**Dataset and training.** Considering the resolution limit of the event camera, a 22cm x 16cm x 17cm size fish tank is chosen to ensure that the camera could capture the aquarium scene clearly and completely. During the experiment, the water depth of the tank is 12cm, resulting in a water volume of approximately 4 liters. Rabbane et al.[17] observed that the optimal breeding density is 5 zebrafish per liter. To simulate a reasonable breeding density for a laboratory environment, 20 zebrafish are put into a 4-liter tank. The process of image acquisition is as follows:

(1) The event camera is aimed directly at a breeding tank containing 20 zebrafish to collect the event stream data.

(2) When the collection is finished, the event stream data is converted into videos in four modes: binary, count, gray, and accumulate. Then, videos of these four modes are fused to generate a mixed mode video, resulting in five videos with the same content.

(3) For these five videos, image frames are extracted every 5 seconds. Since the zebrafish swim randomly, the fixed time interval will not affect the samples' representativeness.

In dataset images acquirement, the whole capturing process lasts for 60 minutes. For each video mode, 720 images are extracted. The LabelMe tool is used for manual labeling. Each target is marked with a bounding box. Zebrafish are labelled as class "zebrafish" and Reflections of zebrafish are labelled as class 'negative sample'. The annotated dataset is completed using images in the mixed mode, which contains 14,042 zebrafish samples and 3512 negative samples. Since the five modes of videos are aligned in time axis, the position of each bounding box is almost the same for the five modes. Local adjustments will be made according to the actual situation The construction process of the dataset is shown in Fig. 4. Images and labels of each dataset are divided into training set and validation set at an 8:2 ratio. The datasets will be put into the YOLOv5 networks to obtain pretrained models.

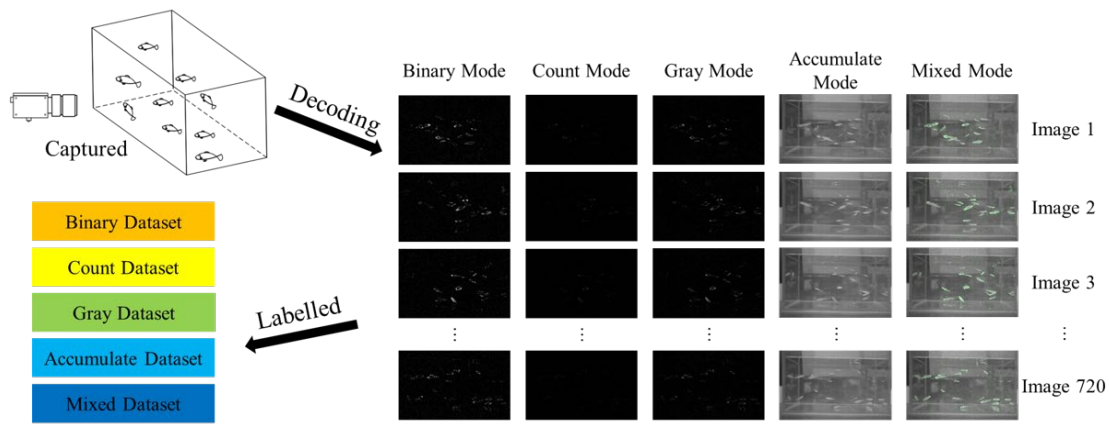

**Figure 4.** Dataset construction process.

## Experiments
**Single image counting.** To valid the detection performance of each model, 8 minutes and 20 seconds event stream data were collected. Then, 100 images of each mode were generated, the extracted method of these images is same as the method to acquire dataset images. These images are isolated from other images. Detection is performed using the pretrained models for each mode separately, and the accuracy comparisons for each model are summarized in Table 1.

Table 1: Accuracy of each mode

| Mode Name | Average Accuracy |
| --- | --- |
| binary | 75% |
| count | 74% |
| gray | 70% |
| accumulate | 81% |
| mixed | 94% |

It can be seen from Table 1 that mixed mode achieves the highest accuracy. However, there are still two situations in which the detection accuracy will be decreased apparently. The one is the occlusion and the other is the mirror artifact. Because dataset has samples for occlusive zebrafish, the pretrained model can detect zebrafish with partial body exposure. With the negative samples labelled, the pretrained model will not regard the mirror reflections as targets. A typical occlusion and mirror artifact situation is shown in result 1 of Fig. 5A. There is still the possibility of missed detections when zebrafish are in a completely overlapping state. One of the completely occlusion situations is in result 2 of Fig. 5A, the completely occluded fish (highlighted in orange box) are enlarged. From the ghost reflection on the water surface, one zebrafish is completely occluded by another and it's been missed during the detection process. When detecting with a single image, duplicate detections can occur, especially in cases where two zebrafish intersect, such as when they form an "X" shape. In such scenarios, the algorithm might mistakenly detect parts of a single zebrafish, as well as its whole body, as separate instances of zebrafish. An example of such a case is shown in Fig. 5B. To mitigate this type of misdetection, zebrafish trajectory information is introduced. By tracking the zebrafish across consecutive video frames, the algorithm can perform a better detection under the circumstances of zebrafish intersection.

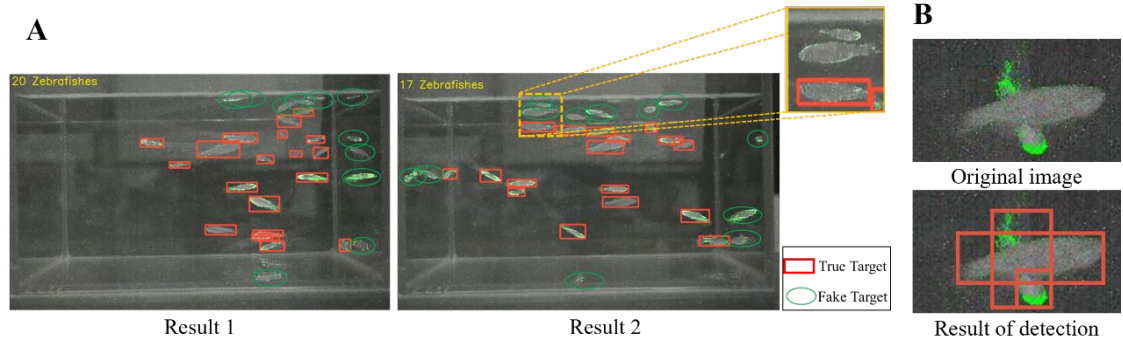

**Figure 5. (A)** Detection results from mixed mode. Reflections of mirror are marked in green circle and they are not detected as targets. **(B)** An instance of duplicate detection.

**Tracking experiment.** The performance of zebrafish tracking is evaluated using the MOTA (Multiple Object Tracking Accuracy) metric in the experiment. The calculation formula for MOTA is shown in Equation (6):

$$MOTA = 1 - \frac{\sum FP + \sum FN + \sum IDS}{\sum GT} \tag{6}$$

where $\sum FP$ represents the number of false positives, $\sum FN$ represents the number if ID switches, $\sum IDS$ represents the number of ID switches, $\sum GT$ represents the total number of targets. The value range of MOTA is $[-\infty, 1]$, with higher values indicating better performance.

In conducting association effect experiments, 10 groups of mixed mode zebrafish videos are generated. Among these 10 groups of video images, group 1-5 were used for tracking performance testing when the zebrafish were not occluded, while group 6-10 were used for tracking performance testing when the zebrafish were occluded. The video frame rate is 30 frames per second. These video images are isolated from other images. The results are shown in Table 2.

Table 2: MOTA accuracy for zebrafish tracking

| Group | Frame Quantity | FP | FN | IDS | GT | MOTA(%) |
|---|---|---|---|---|---|---|
| 1 | 93 | 0 | 0 | 0 | 1860 | 100 |
| 2 | 102 | 0 | 0 | 0 | 2040 | 100 |
| 3 | 110 | 0 | 0 | 0 | 2200 | 100 |
| 4 | 82 | 0 | 0 | 0 | 1640 | 100 |
| 5 | 76 | 0 | 0 | 0 | 1520 | 100 |
| 6 | 33 | 0 | 0 | 0 | 660 | 100 |
| 7 | 25 | 0 | 0 | 0 | 500 | 100 |
| 8 | 45 | 0 | 0 | 97 | 752 | 87.1 |
| 9 | 55 | 0 | 0 | 279 | 960 | 70.9 |
| 10 | 61 | 0 | 0 | 85 | 1105 | 92.3 |
| Average | 68.2 | 0 | 0 | 46.1 | 1323.7 | 95.0 |

In Group 1 to 5, where there is no occlusion of the zebrafish, the algorithm detects and tracks zebrafish correctly. A tracking example without occlusion is shown in the figure Fig. 6A. In Group 6 and 7, the occlusion occurs only in the form of intersecting movement trajectories, without complete occlusion. In these cases, the algorithm still detects each zebrafish accurately, an example of such a case is shown in Fig. 6B. In Group 8 to 10, complete occlusion occurs. There are two types of complete occlusion:

(1) The occluded zebrafish maintains its pre-occlusion motion and reappears within the 15 frames of the video. In this case, the algorithm can accurately track the zebrafish, as it uses Kalman filtering to predict the occluded zebrafish's position. Since the zebrafish's motion remains consistent, when the zebrafish reappears, the algorithm associates it with the same zebrafish from before the occlusion, successfully tracking the zebrafish during short-term occlusion. A typical example of this is when two zebrafish intersect, and one is completely occluded by the other during intersection. Although occlusion occurs, the algorithm can predict the position of the occluded zebrafish using Kalman filtering. An example of such a case is shown in Fig. 6C.

(2) The occluded zebrafish changes its trajectory or does not reappear within 15 frames. Since 15 frames is the maximum number of frames for Kalman filter predictions in our tracking algorithm, if the zebrafish is not detected within this frame count, it is regarded as disappeared. A typical example is when a school of zebrafish gathers. In this case, the school exhibits chaotic movement, leading to more frequent occlusions, and effective tracking becomes impossible. An example of such a case is shown in Fig. 6D.

However, in our experiments, apart from cases where external stimuli (such as a person approaching or feeding) caused the zebrafish to react, the zebrafish did not exhibit clustering behavior. Such occlusion-related tracking issues were not commonly encountered. Though the algorithm did not track some zebrafish precisely, the counting result are correct if there was no complete occlusion.

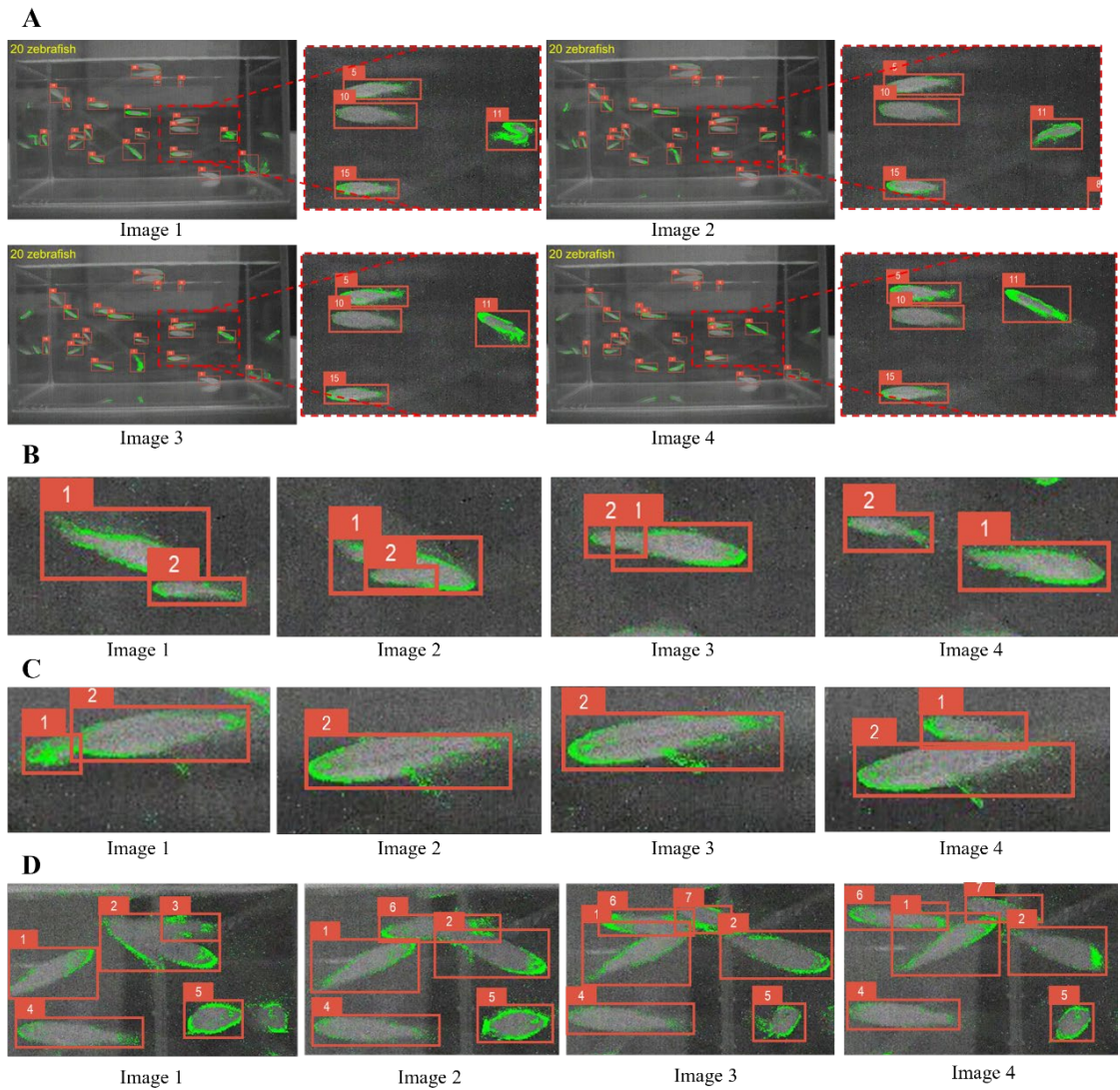

**Figure 6. (A)** When zebrafish are not occluded, all zebrafish are tracked precisely. **(B)** Zebrafish 1 and zebrafish 2 intersect (Not completely occluded), the algorithm can accurately track both. **(C)** Zebrafish 1 is completely occluded by zebrafish 2. After occlusion, zebrafish 1 reappears, and the algorithm successfully associates it with its previous ID before occlusion. **(D)** When zebrafish aggregate as a school, the number of zebrafish can still be correct, but effective tracking becomes more challenging as there are multiple occlusions.

**Multiple images counting.** To validate the accuracy of this statistical mechanism, 100 groups of 3 seconds continuous videos of mixed mode are generated (FPS=30). These videos are isolated in the experiments. The accuracy of these data points were calculated and the results are presented in Fig. 7A. Among them, the detection accuracy with no errors occurred 64 times, with an accuracy of 95% (1 zebrafish error) occurring 31 times, and with an accuracy of 90% (2 zebrafish error) occurring 5 times. The average detection accuracy reaches 97.95%. In image 1 of Fig. 7B, one zebrafish is completely occluded by a zebrafish (ID:17), and the number of zebrafish is 19. In image 2 of Fig. 7B, the occluded zebrafish is detected, and the number of zebrafish becomes 20. After averaging the detection results and rounding up, the final detection count for these two images is 20 zebrafish. In such conditions, the statistical mechanism helps reduce the impact of complete occlusion on the counting results, ensuring more accurate counting in cases where some zebrafish are temporarily occluded.

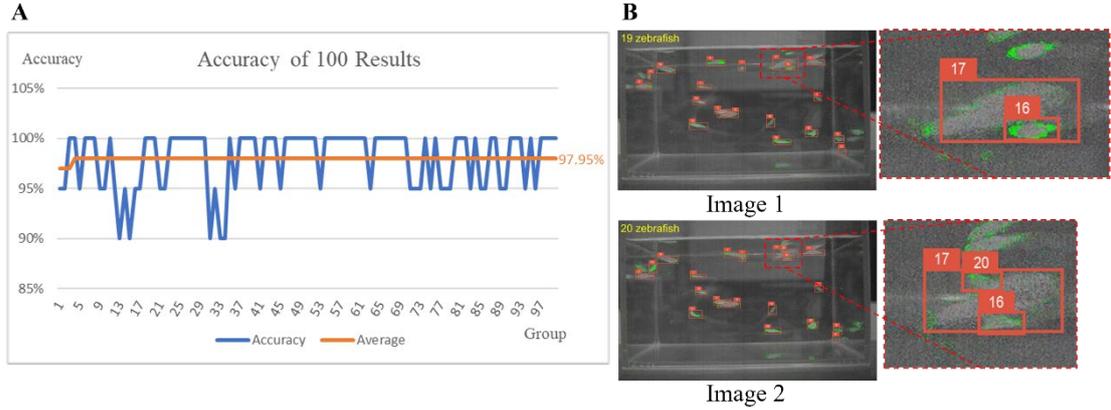

**Figure 7. (A)** Accuracy of 100 sets of 3 seconds continuous randomly selected videos. **(B)** A situation that the statistical mechanism can help to improve counting accuracy.

**Comparison with existing method.** Comparison between existing counting algorithms and ours are given in Table 3.

Table 3: Comparisons of different methods

| Capture method | Algorithm | Effect | Limitations |
| --- | --- | --- | --- |
| Monocular camera[12] | Binarization extraction + threshold setting | Average accuracy of 96.64% | Need single background |
| Monocular camera[7] | GMM + Blob | Average accuracy of 93% | Additional lighting is required |
| Dual underwater camera[8] | ANN | Accuracy > 87% | Make influence for breeding tank |
| Sonar[10] | Edge detection + Kalman filter | Average accuracy of 95% | Sonar is too large for deployment |
| Sonar[9] | Autoregressive time series model | Accuracy > 94% | Sonar is too large for deployment |
| Sonar + underwater camera[11] | Mask RCNN | 68% < Accuracy < 95% | Fit for large fishes |
| Event camera (ours) | Image fusion +denoise+calibration+ YOLO v5+tracking+ statistical mechanism | Average accuracy of 97.95% | |

The binarization extraction plus threshold method achieves a high level of accuracy but requires a simple background. It involves extracting the fish from the background for counting. The GMM background subtraction and blob counting method needs to make the recognized areas fill the whole frame, otherwise the method may subtract incorrectly. Additionally, this method is tested with light supplement lamp, which limits the deployment. Underwater cameras and sonar devices are unsuitable for laboratory breeding tank applications. Our method does not require supplementary lighting during testing, resulting in minimal disturbance to the fish's natural environment. Event cameras capture the contours of moving objects. This character makes our method more adaptable to varying backgrounds and suitable for fish detection in laboratory breeding tanks.

**Conclusion**

In this paper, an event camera-based zebrafish counting algorithm is proposed, which fuses multi-mode images of the event camera, performs tracking and uses a statistical fine-tuning mechanism. The advantage of the statistical mechanism is that it references more detection data and the disadvantage is that accidental errors can amplify the accuracy of the estimation. The cause of accidental error is the gathering behavior of zebrafish, it results in a suddenly decrease of detection number. In the process of application, zebrafish tend to gather sporadically rather than as a whole group, this may only cause the misdetection for one or two targets. And the reasonable breeding density of the breeding tank will not cause large-scale obstruction. Therefore, the results of the mechanism remain stable. Future work will be focused on multiple cameras detection to further improve the detection accuracy. The proposed algorithm can also be applied to the counting of other small organisms.

**Code Availability**

All data, models, or code generated or used during the study are available from the corresponding author by request.

**Data Availability**

The datasets generated and analyzed during the current study are available in the https://drive.google.com/drive/folders/1hY02CWqQ9Jk1oxvtLhNb0LlOMY_gLkF_?usp=sharing.

**Funding:** This work has no founding.

**Competing interests**

The authors declare no competing interests.